\documentclass[runningheads]{llncs}

\usepackage[year=2026]{eccv}

\usepackage{eccvabbrv}
\usepackage{amsmath}
\usepackage{graphicx}
\usepackage{booktabs}
\usepackage{xcolor}
\usepackage{iftex}
\ifPDFTeX
\usepackage[accsupp]{axessibility}
\fi
\usepackage[breaklinks,colorlinks,citecolor=eccvblue]{hyperref}
\usepackage{multirow}
\usepackage{pifont}
\newcommand{\cmark}{\ding{51}}
\providecommand{\CheckmarkBold}{\cmark}
\begin{document}

\title{There and Back Again: A Flexible-Frame Transformer for Multi-Exposure Fusion}
\titlerunning{There and Back Again}
\authorrunning{L. Qu et al.}
\author{Lishen Qu$^{1,3,5}$, Yao Liu$^{1,3,5}$, Shihao Zhou$^{1,3}$ \\ Jie Liang$^{5}$, Hui Zeng$^{5}$, Lei Zhang$^{4,5}$, Jufeng Yang$^{1,2,3,}$\thanks{Corresponding Author.} \\
{$^{1}$Nankai International Advanced Research Institute (SHENZHEN·FUTIAN)} \\
{$^{2}$Peng Cheng Laboratory \qquad $^{3}$College of Computer Science, Nankai University} \\
{$^{4}$The Hong Kong Polytechnic University \qquad $^{5}$OPPO Research Institute}\\
{\tt\small \ \{qulishen, liuyao, zhoushihao96\}@mail.nankai.edu.cn, \ liang27jie@163.com},   
{\tt\small  \ cshzeng@gmail.com,\ cslzhang@comp.polyu.edu.hk, \ yangjufeng@nankai.edu.cn} \ \\
}
\institute{}

\maketitle

\begin{abstract}

Multi-exposure fusion (MEF) brings the dynamic range of conventional cameras closer to that of human vision, producing images with rich scene content. Given the large variability in scene luminance, exposure strategies often require different numbers of frames to capture the full radiance range faithfully. However, conventional MEF techniques are typically designed for a fixed number of inputs, forcing deployment systems to maintain separate models for different frame-count requirements, which undermines deployment efficiency. To address this limitation, we propose FreeMEF, the first flexible-frame transformer for MEF that seamlessly accommodates varying numbers of input exposures without retraining or architectural changes. The proposed approach consists of two key modules. First, we introduce a recurrent state space module (RSSM) that sequentially fuses features from arbitrary sequences via adaptive alignment and state-space recurrent modeling, thereby providing global information guidance for the subsequent restoration. Second, we devise a global feature guided block (GFGB) incorporating an extremity-aware hybrid attention (EAHA) and an affine-injection feed-forward network (AFFN), which effectively resolves the similarity paradox while simultaneously optimizing contrast and brightness regulation. Extensive experiments on three benchmark datasets demonstrate the effectiveness of our method, which performs favorably against state-of-the-art methods both quantitatively and qualitatively. The code is available at \url{https://github.com/qulishen/FreeMEF}.

\end{abstract}

\section{Introduction}
\label{sec:intro}

Conventional imaging systems~\cite{fossum1998digital} are fundamentally constrained by sensor capacity~\cite{sensorcapacity}, making it difficult to capture the full dynamic range of natural scenes.  As a result, high dynamic range (HDR) imaging~\cite{kalantari2017deep,mobilehdr,xu2020u2fusion} plays a vital role in faithfully capturing real-world scenarios with substantial illumination variations. Single-frame methods~\cite{traditional1,traditional2,traditional3} attempt to enhance dynamic range by estimating and adjusting the brightness distribution from the individual image. Nevertheless, such approaches struggle to recover details in regions where information has already been lost due to saturation or underexposure. Therefore, multi-exposure fusion (MEF)~\cite{hasinoff2016burst} has become a widely adopted strategy, capturing multiple low dynamic range (LDR) images at different exposure levels and fusing them to reconstruct an HDR image that preserves more details. Recently, learning-based MEF approaches~\cite{kong2024safnet, AFUNet,liu2022ghost-free_hdrtransformer} have further advanced HDR imaging by leveraging deep neural networks to capture complex lighting conditions, non-linear camera responses, and dynamic scene content.

\begin{figure}[t]\footnotesize
    \centering
    \includegraphics[width=0.9\textwidth]{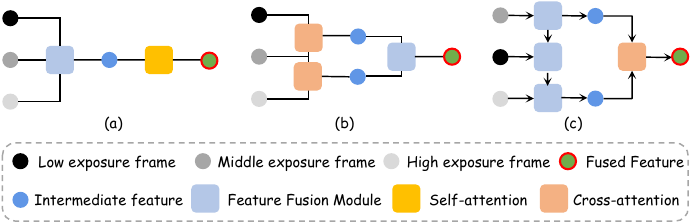}
   \caption{Comparison of feature fusion pipelines. (a) \textbf{Fusion $\rightarrow$ Self-Attention}: Features from three frames are fused prior to self-attention. (b) \textbf{Cross-Attention $\rightarrow$ Fusion}: Reference frames interact with the base frame via cross-attention before fusion. (c) \textbf{Recurrent Fusion $\rightarrow$ Cross-Attention} (Ours): Features are fused sequentially, and the aggregated feature is then incorporated into the base frame via cross-attention.}
    \label{fig:motivation}
\end{figure}

Despite their promising performance, these methods still suffer from two fundamental limitations that hinder practical application. First, existing methods universally assume that the camera uses a fixed exposure strategy. Therefore, these architectures are specifically designed for fixed exposure counts (e.g., 2, 3, or 5) and predefined exposure levels~\cite{chen2025ultrafusion,jiang2023meflut,kong2024safnet}. When the exposure strategy changes, the original model becomes inapplicable, requiring adjustments to the architecture and retraining. Second, many HDR imaging methods~\cite{AFUNet,yan2019attention,mobilehdr} employ attention mechanisms~\cite{vaswani2017attention,liu2021swin} to capture inter-frame dependencies. Some of them~\cite{mobilehdr,liu2024pasta,tel2023alignment_sctnet} perform self-attention after feature fusion (\cref{fig:motivation} (a)), which is prone to introducing ghosting artifacts.  Alternatively, others~\cite{AFUNet,liu2022ghost-free_hdrtransformer,yan2019attention} apply cross-attention between reference and base frames before fusion (\cref{fig:motivation} (b)). While effective for fixed-length inputs, extending such pairwise interaction designs to variable-length exposure sequences is non-trivial and often requires architectural changes or repeated pairwise operations without a unified multi-reference aggregation.  Furthermore, we uncover a limitation in conventional cross-attention, which we define as the \textit{similarity paradox}. Standard cross-attention mechanisms inherently prioritize regions with high inter-frame similarity, which proves effective in tasks with consistent degradation~\cite{Burstormer,burstsr,luo2022bsrt}. However, HDR imaging demands the recovery of saturated (over-exposed) regions that exhibit extremely low pixel-wise similarity to their counterparts in normal-exposure frames~\cite{kong2024safnet}.  This discrepancy hinders the model's ability to effectively aggregate complementary information from neighboring frames. 

To overcome these limitations, we propose FreeMEF, a flexible and robust multi-exposure fusion framework that supports arbitrary input sequence lengths during both training and inference. We reformulate the fusion paradigm by first aggregating multi-frame information into a global representation, and subsequently performing cross-attention with the base frame, as depicted in~\cref{fig:motivation}(c). On the one hand, we propose a recurrent state space module (RSSM), inspired by recurrent structures in video restoration~\cite{turtle,wang2019edvr,kim2018spatio} and understanding~\cite{li2023uniformerv2,huang2018makes,wang2024omnivid} tasks. Instead of constraining the network to a fixed number of inputs, we adopt a shared feature extraction module that processes each exposure frame independently into a latent feature space.  These latent features are integrated via a recurrent mechanism that sequentially updates the aggregated representation with the current frame's information. This design naturally accommodates arbitrary input lengths without requiring any architectural adjustments. On the other hand, we introduce a global feature guided block (GFGB) composed of an extremity-aware hybrid attention (EAHA) and an affine-injection feed-forward network (AFFN).  To mitigate the similarity paradox, the EAHA utilizes a learned extremity map to adaptively balance the contributions of cross-attention and self-attention, ensuring precise feature retrieval. Subsequently, the AFFN refines the fused feature by rectifying brightness and spatial offsets via a gated affine transformation. Anchored to the base frame, these components selectively leverage global context from the RSSM to enhance restoration, thereby significantly reducing motion ghosting.  Consequently, these designs allow FreeMEF to handle arbitrary input numbers with superior exposure fusion performance. In summary, our main contributions are summarized as follows: 
\begin{itemize}
    \item  We propose FreeMEF, a flexible-frame transformer for multi-exposure fusion that conceptually decouples reference features from base features. This design overcomes the constraint of fixed-input architectures, allowing a single model to handle an arbitrary number of input frames during both training and inference without redesign or retraining.
    \item  We propose RSSM, which progressively aggregates exposure features recurrently to accommodate variable input lengths, and GFGB, which exploits global context to guide the base frame, effectively resolving the similarity paradox in saturated regions and achieving precise brightness control.
    \item Comprehensive experiments demonstrate that our FreeMEF achieves superior performance compared to the state-of-the-art methods in both quantitative metrics and visual quality.
\end{itemize}
\section{Related Work}
\label{sec:relatedwork}

\noindent \textbf{HDR Imaging.} HDR imaging methods are broadly divided into two categories: HDR reconstruction~\cite{kalantari2017deep,song2022selective,tel2023alignment_sctnet} and multi-exposure fusion (MEF)~\cite{li2020fast,ma2017robust,xu2020u2fusion}, depending on whether fusion is performed in the linear domain or directly in the LDR domain.  HDR reconstruction methods~\cite{tel2023alignment_sctnet} typically take three bracketed exposures and aim to align all frames to a reference exposure (often the middle one) before merging.  The reconstructed HDR output then requires a separate tone-mapping step~\cite{reinhard2002photographic,durand2002fast} for visualization on standard displays, which may introduce additional artifacts and errors.  In contrast, MEF methods~\cite{jiang2023meflut,chen2025ultrafusion,decomposition} fuse the input exposures directly in the LDR domain, avoiding camera response function (CRF) calibration and the subsequent tone-mapping stage. Considering the above, we choose to conduct the training and comparison of methods on MEF in this work.

\noindent \textbf{Deep Learning-Based Methods.} Early HDR methods~\cite{traditional7,traditional6,traditional5} fused multiple LDR frames through joint optimization, but struggled to handle motion and complex textures. With the rapid development of deep learning~\cite{lecun2015deep,krizhevsky2012imagenet,deng2026towards}, learning-based methods have shown strong performance across a wide range of vision tasks, including detection~\cite{girshick2015fast}, segmentation~\cite{seg1}, and image restoration~\cite{img_restore1}. Kalantari~\etal~\cite{kalantari2017deep} constructed an HDR dataset containing motion and adopted an end-to-end pipeline that first aligns input LDR frames via optical flow~\cite{wu2018deep} and then merges them into an HDR result. Kong~\etal~\cite{kong2024safnet} explicitly learned masks for informative regions, concatenated the three frames, and then performed further refinement. Chen~\etal~\cite{chen2025ultrafusion} treated the MEF task as an inpainting problem, leveraging low-exposure frames to fill in over-exposed regions, which significantly reduces ghosting artifacts in dynamic scenes.

\noindent \textbf{Multi-Frame Feature Fusion.} 
Robust multi-frame feature fusion is crucial for HDR imaging and other burst image tasks~\cite{hasinoff2016burst,burstsr,zhu2024task}. Some methods~\cite{Burstormer,chen2023improving,kong2024safnet} first align frames or features and then fuse them via feature concatenation followed by convolutional layers or attention.  Others~\cite{tel2023alignment_sctnet,liu2022ghost-free_hdrtransformer,chen2022attention} bypass explicit alignment and directly fuse multi-exposure features, allowing the network to implicitly learn local and global motion from data. However, all of these methods assume a fixed number of input frames for the fusion process. In practical MEF applications, different devices, or even different modes of the same device, capture varying numbers of exposure brackets (commonly 2, 3, or 5)~\cite{pourreza2015exposure}. Explicit alignment methods~\cite{AFUNet,kalantari2017deep,kong2024safnet} designed for three frames cannot be trained or evaluated on inputs with 2 or 5 frames, while alignment-free methods~\cite{zamir2022restormer,ASTv2,guo2025mambairv2} still require modifications to the fusion layers (e.g., channel dimensions) and retraining when the number of input frames changes.

\section{Preliminaries}
\label{sec:pre}
\noindent \textbf{Standard State Space Models.}
State space models (SSM)~\cite{gu2024mamba} map a 1D input stimulation $x(t) \in \mathbb{R}$ to an output response $y(t) \in \mathbb{R}$ through a latent state $h(t) \in \mathbb{R}^N$. This continuous system is defined by linear ordinary differential equations (ODEs) characterized by the evolution parameter $\mathbf{A}$, projection parameter $\mathbf{B}$, and output parameter $\mathbf{C}$:
\begin{equation}
h'(t) = \mathbf{A}h(t) + \mathbf{B}x(t), \quad y(t) = \mathbf{C}h(t).
\end{equation}
In modern deep learning implementations like Mamba~\cite{liu2024vmamba,guo2024mambair}, this continuous system is transformed into a discrete form to operate on sampled data. Using the zero-order hold (ZOH) method with a timescale parameter $\mathbf{\Delta}$, the discrete parameters are derived as $\mathbf{\overline{A}} = \exp(\mathbf{\Delta} \mathbf{A})$ and $\mathbf{\overline{B}} = (\mathbf{\Delta} \mathbf{A})^{-1}(\exp(\mathbf{\Delta} \mathbf{A}) - \mathbf{I}) \cdot \mathbf{\Delta} \mathbf{B}$. The discretized recurrence is formulated as:
\begin{equation}
h_t = \mathbf{\overline{A}} h_{t-1} + \mathbf{\overline{B}} x_t, \quad y_t = \mathbf{C} h_t.
\end{equation}
\noindent \textbf{Attentive State-Space Equation.}
While standard SSMs offer linear complexity, the fixed output matrix $\mathbf{C}$ restricts the model to a causal receptive field, potentially missing semantic correlations from unscanned regions. To overcome this, we adopt the attentive state-space equation (ASE)~\cite{guo2025mambairv2}, which modifies the output transformation to facilitate global pixel querying.
The core idea of ASE is to inject instance-specific semantics into the output matrix. This is achieved by constructing a prompt pool $\mathcal{P} \in \mathbb{R}^{N_p \times d}$, where $N_p$ is the number of prompts and $d$ is the hidden state dimension. To ensure parameter efficiency and interpretability, the pool is parameterized via semantic decoupling:
\begin{equation}
\mathcal{P} = \mathbf{M} \mathbf{N},  \quad \mathbf{M} \in \mathbb{R}^{N_p \times r}, \quad \mathbf{N} \in \mathbb{R}^{r \times d},
\end{equation}
where $\mathbf{N}$ is a basis matrix shared across blocks to learn common semantic features, and $\mathbf{M}$ is a block-specific coefficient matrix. $r$ denotes the inner rank such that $r \ll \min\{N_p, d\}$.
To apply these prompts spatially, a routing mechanism selects relevant prompts for each pixel. Given input features, the model predicts a routing probability distribution and employs the Gumbel-Softmax trick~\cite{jang2016categorical} to generate a differentiable one-hot routing matrix $\mathbf{R} \in \mathbb{R}^{L \times N_p}$, where $L$ is the sequence length. The instance-specific prompt matrix $\mathbf{P} \in \mathbb{R}^{L \times d}$ is then obtained by $\mathbf{P} = \mathbf{R} \mathcal{P}$.
Finally, the standard output equation is reformulated to include these learnable prompts, enabling the model to retrieve information from global semantic clusters dynamically:
\begin{equation}
\begin{aligned}
h_{t} &= \mathbf{\overline{A}} h_{t-1}+\mathbf{\overline{{B}}} x_t, \quad y_{t} &= (\mathbf{C} + \mathbf{P}) h_t + \mathbf{D} x_t.
\end{aligned}
\end{equation}
By incorporating $\mathbf{P}$, the ASE mechanism effectively supplements the causal state history with non-causal global context, significantly enhancing the representation capability of the SSM.

\section{Proposed Method}
\begin{figure}[t]\footnotesize
    \centering
    \includegraphics[width=0.95\textwidth]{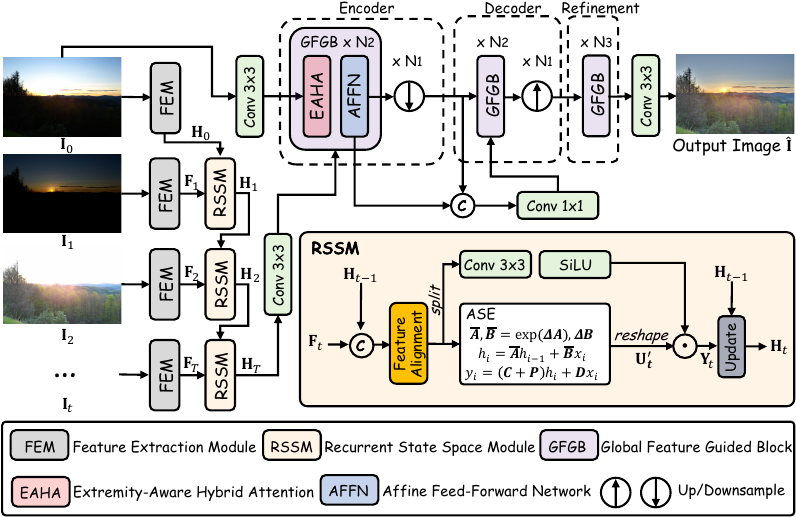}
   \caption{Overview of our FreeMEF pipeline. Unlike previous methods~\cite{AFUNet,kong2024safnet,tel2023alignment_sctnet} that process multiple frames in parallel, we design a recurrent fusion structure based on the ASE~\cite{guo2025mambairv2}, which can adaptively exclude redundant features. Additionally, our model can be transferred to a different number of exposure levels without modification or retraining. The fused global features are then incorporated as auxiliary information to the base frame and fed into the subsequent U-shaped Transformer architecture.}
    \label{fig:model}
\end{figure}

Given a set of LDR frames $\{\mathbf{I}_t\}_{t=0}^{T}$ captured at different exposure values, where $\mathbf{I}_0$ is designated as the base frame and the remaining frames $\{\mathbf{I}_t\}_{t=1}^{T}$ serve as auxiliary frames (i.e., $T$ auxiliary frames and $T{+}1$ frames in total), our objective is to generate a fused ghost-free image $\mathbf{\hat{I}} \in \mathbb{R}^{H \times W \times 3}$ with enhanced dynamic range. Notably, unlike previous approaches~\cite{AFUNet,tel2023alignment_sctnet,kong2024safnet}, the number of auxiliary frames $T$ need not be fixed and can vary across samples during both training and inference. The overall pipeline of our FreeMEF network is depicted in~\cref{fig:model}. First, a recurrent state space module (RSSM) aggregates the features of the base frame and all auxiliary frames into a single global fused feature $\mathbf{H}_T$. Second, a U-shaped Transformer-based architecture takes the base frame and the global fused feature as inputs and progressively refines the base feature at multiple scales using global feature guided blocks (GFGB). Specifically, the GFGB comprises two modules, namely the extremity-aware hybrid attention (EAHA) and the affine feed-forward network (AFFN).

\subsection{Recurrent State Space Module}

We first process the input frame $\mathbf{I}_t$ using the same lightweight feature extraction module (FEM) embedded within the recurrent cell.  The shallow features $\mathbf{F}_t \in \mathbb{R}^{H \times W \times C}$ are obtained as:

\begin{equation}
\mathbf{F}_t = \mathrm{FEM}(\mathbf{I}_t) = W^{2}_{s}\big(\mathrm{ReLU}(W^{1}_{s}\mathbf{I}_{t})\big), \quad t \in \{0,1,\dots,T\},
\end{equation}
where $W^{(\cdot)}_{s}$ denotes a standard $3\times3$ convolution. To effectively model temporal dependencies while handling an arbitrary number of input frames, we design a recurrent fusion scheme. Recently, state space models (SSMs)~\cite{liu2024vmamba,guo2024mambair,guo2025mambairv2} have demonstrated significant potential in modeling long-range dependencies with linear computational complexity. Inspired by this, we incorporate an attentive state-space equation (ASE) in~\cite{guo2025mambairv2} into our recurrent unit to capture global spatial-temporal contexts efficiently.
Let $\mathbf{H}_{t} \in \mathbb{R}^{H \times W \times C}$ denote the hidden state accumulated from the $t$-th frame. We set $\mathbf{H}_{0}$ as an initial tensor which is equal to $\mathbf{F}_0$. For each step $t$, the module updates the hidden state by fusing the current feature $\mathbf{F}_t$ with the history state. First, we employ the deformable alignment strategy~\cite{dai2017deformable} to address the spatial misalignment caused by object motion between the history state and the current frame.  We predict offsets based on the previous hidden state and the current feature, aligning $\mathbf{F}_t$ to $\mathbf{H}_{t-1}$:
\begin{equation}
\begin{aligned}
\bar{\mathbf{F}}_t &= \mathrm{DCN}\big(\mathbf{F}_t, \Delta \mathbf{P}_t\big), \quad \mathrm{where} \quad  \\
\Delta \mathbf{P}_t = \mathcal{F}_{\mathrm{offset}}&\big([\mathbf{H}_{t-1}, \mathbf{F}_{t}]\big)= W^{4}_s\mathrm{LeakyReLU}(W^{3}_s[\mathbf{H}_{t-1},\mathbf{F}_{t}]),  
\end{aligned}
\end{equation}
where $\mathrm{DCN}(\cdot)$ is the deformable convolution. $[\cdot, \cdot]$ denotes concatenation along the channel dimension, and $\bar{\mathbf{F}}_t \in \mathbb{R}^{H \times W \times C}$ represents the aligned current feature. Then, the global features are captured using the ASE, and the output is obtained as $\mathbf{Y}_t$. Specifically, the input feature is projected into a higher-dimensional space and split into a data branch and a gate branch. This process can be represented by:
\begin{equation}
\begin{aligned}
    \mathbf{U}_t, \mathbf{Z}_t = \mathrm{Split}&(\mathrm{Linear}(\bar{\mathbf{F}}_t)), \quad
    \mathbf{U}'_t = \mathrm{ASE}\big(\mathrm{SiLU}(W^{2}_{d}(\mathbf{U}_t))\big), \\
    \mathbf{Y}_t &= \mathbf{U}'_t \odot \mathrm{SiLU}(W^{1}_{d}(\mathbf{Z}_t)),
    \end{aligned}
\end{equation}
where $W^{(\cdot)}_{d}$ is the $3 \times 3$ depth-wise convolution. 
Then, we reshape $\mathbf{Y}'_t \in \mathbb{R}^{HW \times C}$ into $\mathbf{Y}_t \in \mathbb{R}^{H \times W \times C}$.
Finally, the hidden state is updated by fusing $\mathbf{Y}_{t}$ with the previous state $\mathbf{H}_{t-1}$ via a sigmoid activation gate $\mathbf{G}_t = \sigma(W^5_s[\mathbf{Y}_t, \mathbf{H}_{t-1}])$, formulated as:
\begin{equation}
    \mathbf{H}_t = \mathbf{H}_{t-1} + \mathbf{G}_t \odot (W^1_{p}([\mathbf{Y}_{t}, \mathbf{H}_{t-1}]) - \mathbf{H}_{t-1}),
\end{equation} 
where $\sigma(\cdot)$ is the sigmoid function and $W^{(\cdot)}_p$ denotes the $1\times1$ point-wise convolution.

\begin{figure}[t]\footnotesize
    \centering
    \includegraphics[width=\textwidth]{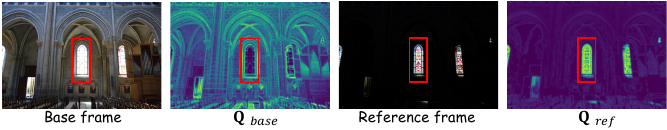}
    \caption{The visualization of the \textit{query} $\mathbf{Q}_{base}$ from the base feature and the \textit{query} $\mathbf{Q}_{ref}$ from the reference feature. In the over-exposed \textcolor{red}{red}-boxed region, the values of $\mathbf{Q}_{base}$ are small, which leads to low attention map values when querying $\mathbf{K}$ in this region. As a result, $\mathbf{Q}_{ref}$ is required to provide supplementary information.}
    \label{fig:similarity}
\end{figure}

\subsection{Global Feature Guided Block}
\label{sec:similarity}

\begin{figure}[t]\footnotesize
    \centering
    \includegraphics[width=\textwidth]{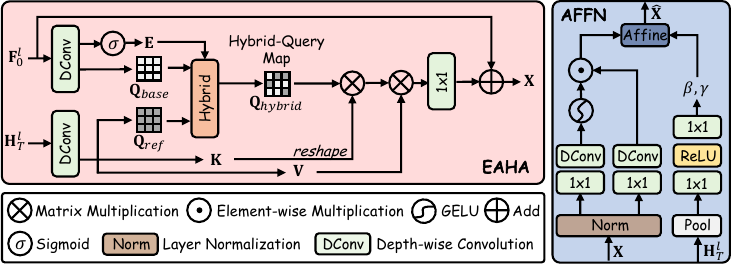}
       \caption{Illustration of GFGB. It comprises an extremity-aware hybrid attention (EAHA) and an affine-injection feed-forward network (AFFN).}
    \label{fig:model2}
\end{figure}

\noindent\textbf{The Similarity Paradox.} Although the global feature $\mathbf{H}_T$ provides information about the reference frames, utilizing it to restore the base frame $\mathbf{I}_{0}$ introduces a fundamental conflict. The primary objective of HDR imaging is to hallucinate details in regions where the base frame is clipped (over-exposed) or noisy (under-exposed)~\cite{kong2024safnet,tel2023alignment_sctnet,chen2025ultrafusion}.  However, standard attention mechanisms are based on similarity matching ($\mathbf{Q}$ vs. $\mathbf{K}$), where $\mathbf{Q}$ is extracted from the base features, and $\mathbf{K}$ is typically derived from the reference features. In over/under-exposed regions, the base features are numerically extreme and structurally degraded, leading to low correlation with the healthy global features, as shown in~\cref{fig:similarity}. Therefore, this leads to a similarity paradox: the attention mechanism assigns low weights to the regions that most require restoration. To resolve this, we propose the global feature guided block (GFGB), as shown in~\cref{fig:model2}. It incorporates an extremity-aware hybrid attention (EAHA) that calibrates the query vectors based on the severity of saturation, ensuring robust feature query even when appearance similarity is low. Besides, we introduce affine transformations into the feed-forward network (FFN), and propose the affine-injection feed-forward network (AFFN), which leverages global features to adjust the brightness and contrast of the base frame.

\noindent\textbf{Extremity-Aware Hybrid Attention (EAHA).}
Let $\mathbf{F}_{0}^{l}, \mathbf{H}^{l}_T \in \mathbb{R}^{\frac{H}{2^l} \times \frac{W}{2^l} \times 2^lC}$ denote the base frame feature and the global feature of the $l$-th stage, respectively. First, we generate the \textit{key} $\mathbf{K}$ and \textit{value} $\mathbf{V}$ from the global feature, and a primary \textit{query} $\mathbf{Q}_{base}$ from the base frame:
\begin{equation}
    \mathbf{Q}_{base} = W^{Q}_dW^{Q}_p\mathbf{F}_{0}^{l}, \quad \mathbf{K} = W^{K}_dW^{K}_p\mathbf{H}^{l}_T, \quad \mathbf{V} = W^{V}_dW^{V}_p\mathbf{H}^{l}_T. \quad 
\end{equation}
Since $\mathbf{Q}_{base}$ is unreliable in saturated areas, we generate a reference \textit{query} $\mathbf{Q}_{ref}$ derived directly from the global feature. This acts as a fallback search vector, allowing the model to perform self-correlation within the history when the base frame fails:
\begin{equation}
    \mathbf{Q}_{ref} = W^{ref}_{d}W^{ref}_{q}\mathbf{H}^{l}_T.
\end{equation}
To blend the two queries, we estimate an extremity map $\mathbf{E} \in \mathbb{R}^{\frac{H}{2^l} \times \frac{W}{2^l} \times 1}$ that highlights saturated pixels, where $\mathbf{E} = \sigma(W^{E}_d\mathbf{F}_0^{l})$.  We then synthesize a hybrid query $\mathbf{Q}_{hybrid}$ and the map $\mathbf{E}$ serves as a spatial gate:
\begin{equation}
    \mathbf{Q}_{hybrid} = (1 - \mathbf{E}) \odot \mathbf{Q}_{base} + \mathbf{E} \odot \mathbf{Q}_{ref}.
\end{equation}
Here, $\odot$ denotes broadcasting element-wise multiplication. In well-exposed regions ($\mathbf{E} \approx 0$), the model queries based on the base frame's content. In saturated regions ($\mathbf{E} \approx 1$), the model switches to $\mathbf{Q}_{ref}$, effectively utilizing the history's own structural priors to find relevant features. Finally, we compute the cross-attention along the channel dimension:
\begin{equation}
    \mathrm{Attention}(\hat{\mathbf{Q}}_{\text{hybrid}}, \hat{\mathbf{K}}, \hat{\mathbf{V}}) = \hat{\mathbf{V}} \mathrm{Softmax}\left(\frac{\hat{\mathbf{K}}^\top \hat{\mathbf{Q}}_{\text{hybrid}}}{\alpha}\right),
\end{equation}
where $\hat{\mathbf{Q}}_{\text{hybrid}}, \hat{\mathbf{K}}, \hat{\mathbf{V}} \in \mathbb{R}^{\frac{HW}{2^l} \times 2^lC}$ are obtained by flattening the original tensors.

$\alpha$ is a learnable temperature scaling parameter. The resulting feature map captures global radiance information while being robust to local clipping.  Essentially, $\hat{\mathbf{K}}^\top \hat{\mathbf{Q}}_{\text{hybrid}}$ is equivalent to a weighted sum of the self-attention map and the cross-attention map without additional computational overhead.

\noindent\textbf{Affine-Injection Feed-Forward Network (AFFN).} Recovering HDR content involves not only simple feature addition, but also addressing the domain shift between LDR and HDR by rectifying the luminance statistics (i.e., scale and shift) of the features~\cite{affine1,affine2}. Standard FFNs lack the ability to dynamically adapt to these global exposure changes. We propose to inject global radiance information into the FFN via affine modulation. We extract a global descriptor vector from the history $\mathbf{H}^l_T$ and predict affine parameters $\gamma, \beta \in \mathbb{R}^{1 \times 1 \times C}$:
\begin{equation}
    \gamma, \beta = \mathrm{Split}(\mathrm{Conv}(\mathrm{GlobalAvgPool}(\mathbf{H}^l_T))).
\end{equation}
Given an input tensor $\mathbf{X} \in \mathbb{R}^{H \times W \times C}$, AFFN is formulated as:
\begin{equation}
\begin{aligned}
    \mathbf{\hat{X}} &= W^2_p \mathrm{Gating}(\mathbf{X}\odot(1+\gamma) + \beta), \quad \\
    \mathrm{Gating}(\mathbf{X}) &= \mathrm{GELU}(W^3_dW^3_p(\mathrm{LN}(\mathbf{X}))) \odot W^4_dW^4_p(\mathrm{LN}(\mathbf{X})).
\end{aligned}
\end{equation}
The $\mathrm{LN}(\cdot)$ denotes the layer normalization~\cite{ba2016layer}. By injecting $(\gamma, \beta)$, AFFN explicitly recalibrates the contrast and brightness of the local features based on the high-dynamic-range statistics preserved in the global history.

\begin{table*}[!t]\footnotesize

    \caption{Quantitative experimental results of training and inference with 3 frames on Kalantari~\etal's dataset~\cite{kalantari2017deep} and Real-HDRV dataset~\cite{shu2024towards_realhdrv}. The best result is on \textbf{bold}, and the second best is \underline{underlined}. }
    \label{table:kalantari} 
    \centering
    \resizebox{\linewidth}{!}{
\begin{tabular}{c|ccc|ccc|cc}
\toprule
\multirow{2}{*}{Method} & \multicolumn{3}{c|}{Kalantari~\etal~\cite{kalantari2017deep}} & \multicolumn{3}{c|}{Real-HDRV~\cite{shu2024towards_realhdrv}} & \multirow{2}{*}{\begin{tabular}[c]{@{}c@{}}Flops\\ (G)\end{tabular}} & \multirow{2}{*}{\begin{tabular}[c]{@{}c@{}}Params\\ (M)\end{tabular}} \\ \cline{2-7}
 & PSNR $\uparrow$ & SSIM $\uparrow$ & LPIPS $\downarrow$ & PSNR $\uparrow$ & SSIM $\uparrow$ & LPIPS $\downarrow$ &  &  \\ \midrule
HDR-Transformer~\cite{liu2022ghost-free_hdrtransformer} (ECCV'22) & 27.024 & \underline{0.935} & 0.093 & 25.131 & \underline{0.930} & \underline{0.079} & 95.261 & 1.454 \\
Restormer~\cite{zamir2022restormer} (CVPR'22) & 27.149 & 0.930 & 0.097 & \underline{25.562} & 0.924 & 0.089 & 141.160 & 26.099 \\
MEFLUT~\cite{jiang2023meflut} (ICCV'23) & 19.362 & 0.747 & 0.270 & 16.392 & 0.655 & 0.405 & 33.060 & 0.356 \\
SCTNet~\cite{tel2023alignment_sctnet} (ICCV'23) & 27.101 & 0.921 & 0.103 & 24.821 & 0.918 & 0.099 & 218.885 & 3.326 \\
SAFNet~\cite{kong2024safnet} (ECCV'24) & 24.284 & 0.927 & 0.094 & 21.467 & 0.908 & 0.111 & 42.589 & 1.121 \\
ASTv2~\cite{ASTv2} (TPAMI'25) & 26.831 & 0.923 & 0.100 & 25.110 & 0.922 & 0.088 & 110.626 & 7.751 \\
MambaIRv2~\cite{guo2025mambairv2} (CVPR'25) & 26.910 & 0.931 & 0.094 & 25.057 & 0.913 & 0.085 & 83.506 & 1.211 \\
AFUNet~\cite{AFUNet} (ICCV'25) & \underline{27.226} & 0.925 & \underline{0.091} & 25.423 & 0.912 & 0.084 & 75.340 & 1.138 \\ \midrule
FreeMEF (Ours) & \textbf{28.418} & \textbf{0.948} & \textbf{0.081} & \textbf{26.077} & \textbf{0.940} & \textbf{0.074} & 41.496 & 8.900 \\ \bottomrule
\end{tabular}
}
\end{table*}

\begin{figure}[t]\footnotesize
    \centering
    \includegraphics[width=\textwidth]{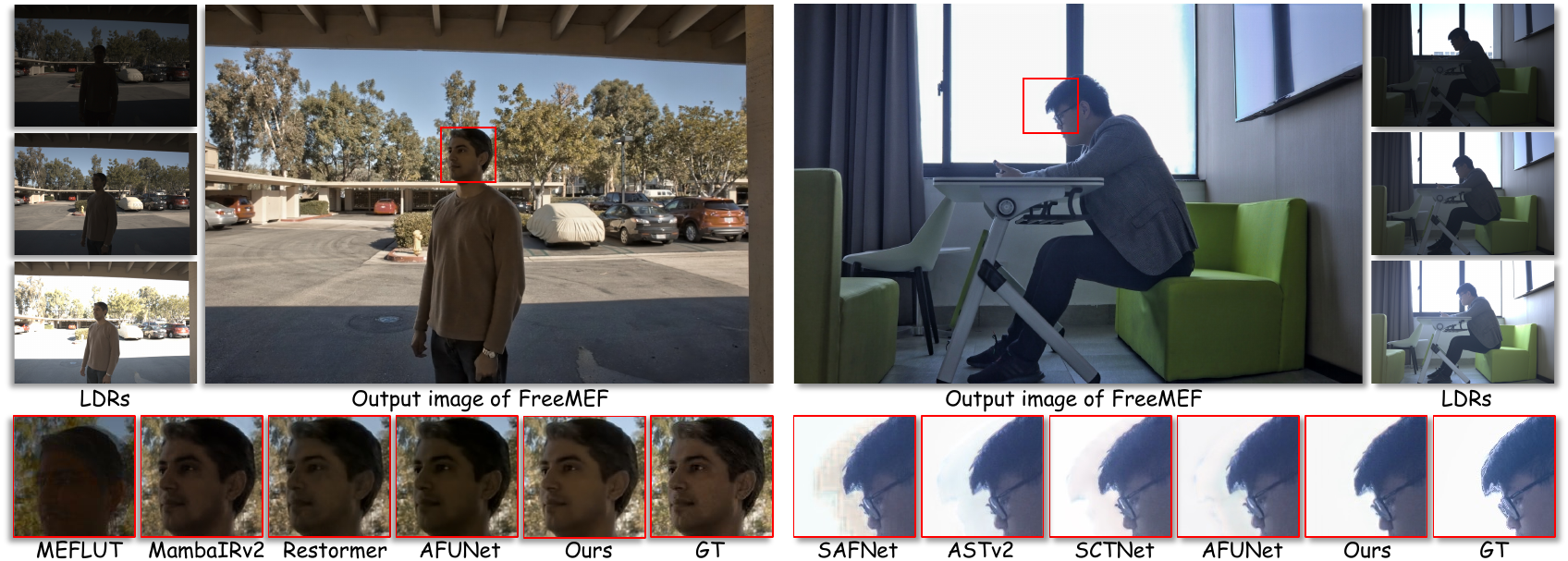}
    \caption{Visual comparisons on Kalantari~\etal's dataset~\cite{kalantari2017deep} (left) and Real-HDRV dataset~\cite{shu2024towards_realhdrv} (right). Our method achieves robust performance under both under-exposed and over-exposed conditions.}
    \label{fig:kalantari}
\end{figure}

\begin{table*}[t]\footnotesize

    \caption{Quantitative results of models trained on Kalantari~\etal's dataset~\cite{kalantari2017deep} and evaluated on different exposure frames of SICE~\cite{jiang2023meflut}. Note that previous models require modifications to their architectures and must be retrained separately on training sets corresponding to each frame number. In contrast, our FreeMEF requires no changes and can be directly tested with different frames.}
    \label{tab:sice} 
    \centering
\resizebox{\linewidth}{!}{
\begin{tabular}{c|ccc|ccc|ccc}
\toprule
\multirow{2}{*}{Method} & \multicolumn{3}{c|}{Test on SICE (2-frame)} & \multicolumn{3}{c|}{Test on SICE (3-frame)} & \multicolumn{3}{c}{Test on SICE (5-frame)} \\ \cline{2-10} 
 & PSNR $\uparrow$ & SSIM $\uparrow$ & LPIPS $\downarrow$ & PSNR $\uparrow$ & SSIM $\uparrow$ & LPIPS $\downarrow$ & PSNR $\uparrow$ & SSIM $\uparrow$ & LPIPS $\downarrow$ \\ \midrule
 HDR-Trans.~\cite{liu2022ghost-free_hdrtransformer} (ECCV'22)& \underline{15.339} & \underline{0.690} & \underline{0.262} & 18.151 & 0.746 & 0.227 & 20.748 & 0.830 & 0.171 \\
 Restormer~\cite{zamir2022restormer} (CVPR'22)& 14.492 & 0.674 & 0.267 & 17.880 & 0.734 & 0.228 & 21.091 & 0.837 & 0.167 \\
MEFLUT~\cite{jiang2023meflut} (ICCV'23)& 13.415 & 0.535 & 0.301 & 15.747 & 0.621 & 0.298 & 16.214 & 0.645 & 0.280 \\
SCTNet~\cite{tel2023alignment_sctnet} (ICCV'23)& 15.288 & 0.685 & 0.264 & \underline{18.724} & \underline{0.768} & \underline{0.210} & \underline{21.153} & \underline{0.842} & 0.177 \\
SAFNet~\cite{kong2024safnet} (ECCV'24)& 15.161 & 0.627 & 0.277 & 16.947 & 0.738 & 0.223 & 18.798 & 0.820 & 0.170 \\
ASTv2~\cite{ASTv2} (TPAMI'25)& 14.003 & 0.646 & 0.286 & 18.240 & 0.734 & 0.227 & 20.971 & 0.820 & 0.185 \\ 
MambaIRv2~\cite{guo2025mambairv2} (CVPR'25)& 14.885 & 0.686 & 0.273 & 18.013 & 0.747 & 0.218 & 20.414 & 0.825 & 0.166 \\
AFUNet~\cite{AFUNet} (ICCV'25)& 14.591 & 0.647 & 0.290 &  18.669 & 0.758 & 0.218 & 21.121 & 0.838 & \underline{0.164} \\
\midrule
FreeMEF (Ours) & \textbf{17.087} & \textbf{0.731} & \textbf{0.225} & \textbf{19.326} & \textbf{0.774} & \textbf{0.199} & \textbf{22.269} & \textbf{0.851} & \textbf{0.149} \\ \bottomrule
\end{tabular}
}
\end{table*}

\begin{figure}[t]\tiny
    \centering
    \includegraphics[width=\textwidth]{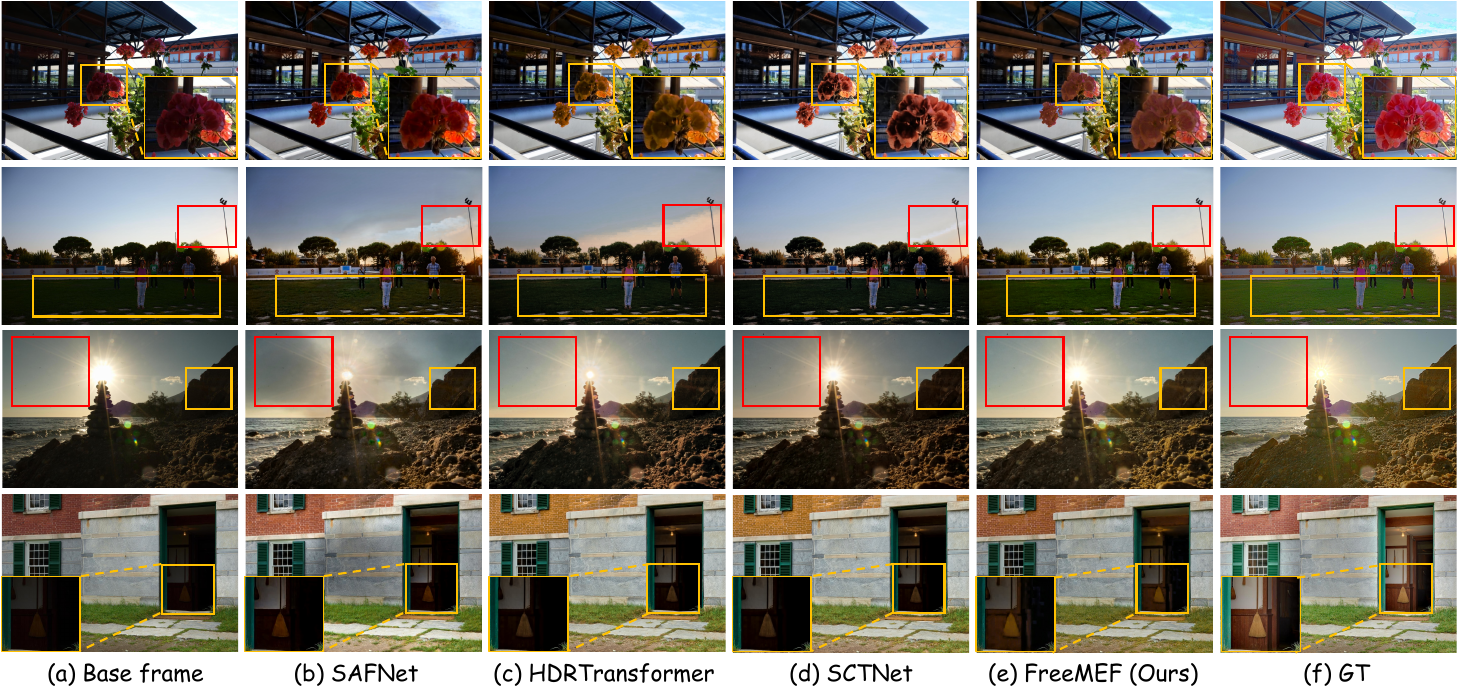}
    \caption{Visual comparisons between models trained on Kalantari~\etal's dataset~\cite{kalantari2017deep} and tested on the SICE dataset~\cite{sice_cai2018learning}. Under-exposed and over-exposed regions are marked with \textcolor[RGB]{255,192,0}{yellow} and \textcolor{red}{red} boxes, respectively. Our method achieves a higher dynamic range and consistently delivers superior visual quality. Zoom in for a better view. Note that Kalantari~\etal's training dataset~\cite{kalantari2017deep} contains fewer exposure levels than SICE~\cite{sice_cai2018learning}, resulting in a lower dynamic range of all model outputs compared to the ground truth.}
    \label{fig:compa1}
\end{figure}

\section{Experiments}
\subsection{Experimental settings}
\noindent\textbf{Datasets.} We use Kalantari~\etal's dataset~\cite{kalantari2017deep} and the Real-HDRV dataset~\cite{shu2024towards_realhdrv} for training, in which each sample consists of three LDR frames and one HDR file.  Following the approach in MEFLUT~\cite{jiang2023meflut}, the HDR files are tone-mapped and converted to PNG format for use in the MEF task. In addition, we conduct cross-dataset evaluation by training on Kalantari~\etal's dataset~\cite{kalantari2017deep} and testing on the SICE dataset~\cite{sice_cai2018learning}. Specifically, since SICE includes multiple exposure levels, we evaluate on its 2-, 3-, and 5-exposure subsets to assess generalization across different frame numbers.

\noindent\textbf{Baselines.}
We select the state-of-the-art HDR imaging methods for comparison, including MEFLUT~\cite{jiang2023meflut}, SAFNet~\cite{kong2024safnet}, SCTNet~\cite{tel2023alignment_sctnet}, HDRTransformer~\cite{liu2022ghost-free_hdrtransformer}, and AFUNet~\cite{AFUNet}.  Since these methods are specifically tailored for three-frame inputs, we minimally adapt their input embedding layers for 2-frame and 5-frame settings to match the expected input dimensionality, while keeping the remaining architectures unchanged. Furthermore, we include representative models from the general image restoration field, such as Restormer~\cite{zamir2022restormer}, MambaIRv2~\cite{guo2025mambairv2}, and ASTv2~\cite{ASTv2}.  For fair comparison, these models are retrained to accommodate 2-frame and 5-frame inputs by adjusting the channel dimensions of their initial embedding layers. To train the 5-frame models, we expand Kalantari~\etal's dataset~\cite{kalantari2017deep} from 3 to 5 frames through exposure simulation~\cite{afifi2021learning}. 

\noindent\textbf{Implementation Details.}
In our default setting, FreeMEF adopts a four-stage encoder-decoder architecture with an additional bottleneck stage.
The channel dimension is set to $dim=32$, and the numbers of blocks in the four stages are configured as $[2,2,2,2]$, followed by $2$ refinement blocks.
During training, we crop image patches of size $256\times256$ and use a batch size of $2$ per GPU, with geometric augmentations enabled.
The model is optimized with the Adam optimizer, where the initial learning rate is set to $2\times10^{-4}$, the weight decay is $0$, and the momentum parameters are $\beta_1=0.9$ and $\beta_2=0.999$.
The learning rate is adjusted by a Cosine Annealing Restart Cyclic scheduler for a total of $300$k iterations, consisting of two periods of $92$k and $208$k iterations, with minimum learning rates of $2.85\times10^{-4}$ and $1\times10^{-6}$, respectively.
Finally, the network is trained using the standard $L_1$ loss.

\subsection{Comparison with Previous Work}
We compare FreeMEF with 8 state-of-the-art methods, as shown in~\cref{table:kalantari}.  On Kalantari~\etal's dataset~\cite{kalantari2017deep}, our method achieves a PSNR that is 1.192 dB higher than the second-best method, and on the Real-HDRV dataset~\cite{shu2024towards_realhdrv}, the improvement is 0.515 dB.  Furthermore, as depicted in~\cref{fig:kalantari}, our method demonstrates superior HDR performance. In the left image, our approach achieves significantly better restoration of backlit faces compared to previous methods. In the right image, previous methods introduce motion artifacts (e.g., ghosting) in the presence of moving subjects against over-exposed backgrounds, whereas our method effectively avoids such artifacts. Our method excels at maintaining low FLOPs, saving nearly 39\% and 56\% in computational cost relative to AFUNet~\cite{AFUNet} and HDR-Transformer~\cite{liu2022ghost-free_hdrtransformer}, respectively. The number of parameters also falls within the acceptable range for deployment on mobile devices~\cite{gao2024quantnas}. 

To further compare the models' robustness, we conduct a cross-dataset evaluation.  Specifically, all methods are trained on Kalantari~\etal's dataset~\cite{kalantari2017deep}, and subsequently tested on the SICE dataset~\cite{sice_cai2018learning}, which consists of exposure sequences with varying numbers of frames.  Accordingly, we evaluate the models on three different SICE test subsets, each with a distinct number of LDR frames. The test results are shown in~\cref{tab:sice}, where our FreeMEF outperforms the second-best method in terms of PSNR by 1.748, 0.657, and 1.116 dB for 2-frame, 3-frame, and 5-frame fusion, respectively. Our method demonstrates particularly significant improvements, especially in 2-frame and 5-frame fusion scenarios, which indicates better generalization to varying numbers of input frames at inference time. 

We attribute this limitation to the two-frame training setting, where existing models are exposed only to long- and short-exposure images and thus fail to acquire prior knowledge of intermediate exposures. Conversely, FreeMEF benefits from training on multi-frame data, enabling it to generalize and exploit rich priors even during 2-frame inference. Regarding 5-frame fusion, fixed-input methods find it more challenging to compress all frame features into the base frame at the beginning of the network, resulting in greater information loss compared to 3-frame fusion. Our recurrent fusion architecture circumvents this bottleneck by processing features sequentially and decoupling the reference features from the base features, thereby preserving detail and significantly enhancing performance. Besides, we present visual comparison results on the highly challenging scenes of the SICE dataset~\cite{sice_cai2018learning}, as shown in~\cref{fig:compa1}. As shown in rows 1–3 of~\cref{fig:compa1}, our FreeMEF demonstrates superior recovery ability and achieves a higher dynamic range, which can be attributed to the synergistic effect of RSSM and EAHA. Furthermore, owing to the introduction of affine modulation in AFFN, FreeMEF achieves better contrast and color fidelity, as illustrated in row 4 of~\cref{fig:compa1}.

\begin{table*}[t]
    \centering
    \caption{Ablation studies of different components. The CNN, MDTA, and GDFN modules are drawn from Restormer~\cite{zamir2022restormer}, while the RSSM, EAHA, and AFFN modules are tailored for our FreeMEF. To ensure fairness in comparison, the parameter budgets of all modules are balanced.}
    \label{tab:ablation}
\resizebox{\linewidth}{!}{
\begin{tabular}{cclcl|clcl|clcl|clcc|cc}
\toprule
    & \multicolumn{2}{c}{CNN}    & \multicolumn{2}{c|}{RSSM}   & \multicolumn{2}{c}{MDTA}   & \multicolumn{2}{c|}{EAHA}   & \multicolumn{2}{c}{GDFN}   & \multicolumn{2}{c|}{AFFN}   & \multicolumn{2}{c}{PSNR $\uparrow$} & SSIM $\uparrow$ & LPIPS $\downarrow$ & Flops (G) & Params (M) \\ \midrule (a) & \multicolumn{2}{c}{\CheckmarkBold} & \multicolumn{2}{c|}{}       & \multicolumn{2}{c}{\CheckmarkBold} & \multicolumn{2}{c|}{}  & \multicolumn{2}{c}{\CheckmarkBold} & \multicolumn{2}{c|}{}       & \multicolumn{2}{c}{27.082}& 0.920 & 0.103&    38.454       &     8.621       \\

(b) & \multicolumn{2}{c}{}       & \multicolumn{2}{c|}{\CheckmarkBold} & \multicolumn{2}{c}{\CheckmarkBold} & \multicolumn{2}{c|}{}       & \multicolumn{2}{c}{\CheckmarkBold} & \multicolumn{2}{c|}{}       & \multicolumn{2}{c}{27.565}& 0.937 & 0.090 &    39.315       &      8.633     \\

(c) & \multicolumn{2}{c}{\CheckmarkBold} & \multicolumn{2}{c|}{}       & \multicolumn{2}{c}{}       & \multicolumn{2}{c|}{\CheckmarkBold} & \multicolumn{2}{c}{\CheckmarkBold} & \multicolumn{2}{c|}{}       & \multicolumn{2}{c}{27.457}& 0.934 & 0.094&     40.872      &   8.801       \\

(d) & \multicolumn{2}{c}{\CheckmarkBold} & \multicolumn{2}{c|}{}       & \multicolumn{2}{c}{\CheckmarkBold} & \multicolumn{2}{c|}{}       & \multicolumn{2}{c}{}       & \multicolumn{2}{c|}{\CheckmarkBold} & \multicolumn{2}{c}{27.426}& 0.928 & 0.085&       38.236    &   8.632      \\ 

\midrule
(e) & \multicolumn{2}{c}{\CheckmarkBold} & \multicolumn{2}{c|}{}       & \multicolumn{2}{c}{} & \multicolumn{2}{c|}{\CheckmarkBold}  & \multicolumn{2}{c}{} & \multicolumn{2}{c|}{\CheckmarkBold}       & \multicolumn{2}{c}{27.982}& 0.937 & 0.095 &    40.615       &    8.893      \\
(f) & \multicolumn{2}{c}{} & \multicolumn{2}{c|}{\CheckmarkBold}       & \multicolumn{2}{c}{\CheckmarkBold} & \multicolumn{2}{c|}{}  & \multicolumn{2}{c}{} & \multicolumn{2}{c|}{\CheckmarkBold}       & \multicolumn{2}{c}{27.837}& 0.938 & 0.089 &    39.117       &    8.639      \\
(g) & \multicolumn{2}{c}{} & \multicolumn{2}{c|}{\CheckmarkBold}       & \multicolumn{2}{c}{} & \multicolumn{2}{c|}{\CheckmarkBold}  & \multicolumn{2}{c}{\CheckmarkBold} & \multicolumn{2}{c|}{}       & \multicolumn{2}{c}{28.073}& 0.942 & 0.084 &   41.761      &    8.826      \\

\midrule (h) & \multicolumn{2}{c}{}       & \multicolumn{2}{c|}{\CheckmarkBold} & \multicolumn{2}{c}{}       & \multicolumn{2}{c|}{\CheckmarkBold} & \multicolumn{2}{c}{}       & \multicolumn{2}{c|}{\CheckmarkBold} & \multicolumn{2}{c}{\textbf{28.418}}& \textbf{0.948} & \textbf{0.081}&     41.496   &     8.900        \\ \bottomrule
\end{tabular}
}
\end{table*}

\begin{figure}[t]\tiny
    \centering
    \includegraphics[width=0.98\textwidth]{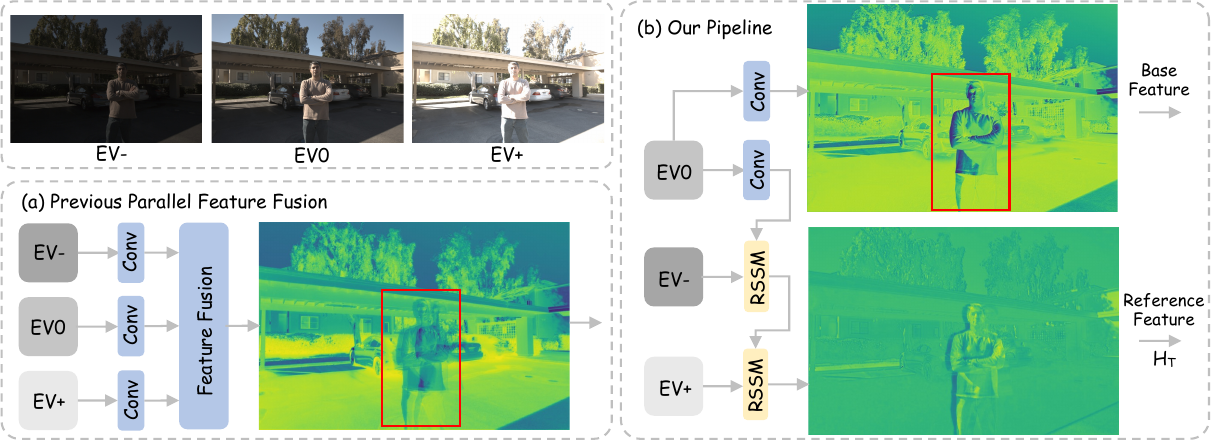}
    \caption{Feature fusion strategy comparison. (a) The parallel fusion strategy treats all frame features equally, which can easily introduce ghosting artifacts. (b) Our method decouples the base frame from the reference features and employs a cross-attention mechanism for adaptive fusion.  Unlike prior parallel fusion methods, our proposed "there and back again" mechanism effectively suppresses ghosting artifacts. 
       }
    \label{fig:feature_fusion}
\end{figure}
\subsection{Analysis and Discussion}
\noindent\textbf{Effect of Individual Modules.} For the ablation studies, we select three modules from Restormer~\cite{zamir2022restormer} for comparison.  Specifically, we replace the RSSM, EAHA, and AFFN modules in our FreeMEF with Restormer's CNN embedding layer, MDTA, and GDFN, respectively, to validate the effectiveness of our proposed designs. As shown in~\cref{tab:ablation} (a)-(d), RSSM, EAHA, and AFFN contribute individually to PSNR gains of 0.472 dB, 0.372 dB, and 0.347 dB, respectively. Besides, for the comparative experiments involving the removal of specific modules, i.e., (e)-(h), removing RSSM, EAHA, and AFFN results in performance drops of 0.526 dB, 0.581 dB, and 0.397 dB, respectively.

\begin{figure}[t]\tiny
    \centering
    \includegraphics[width=\textwidth]{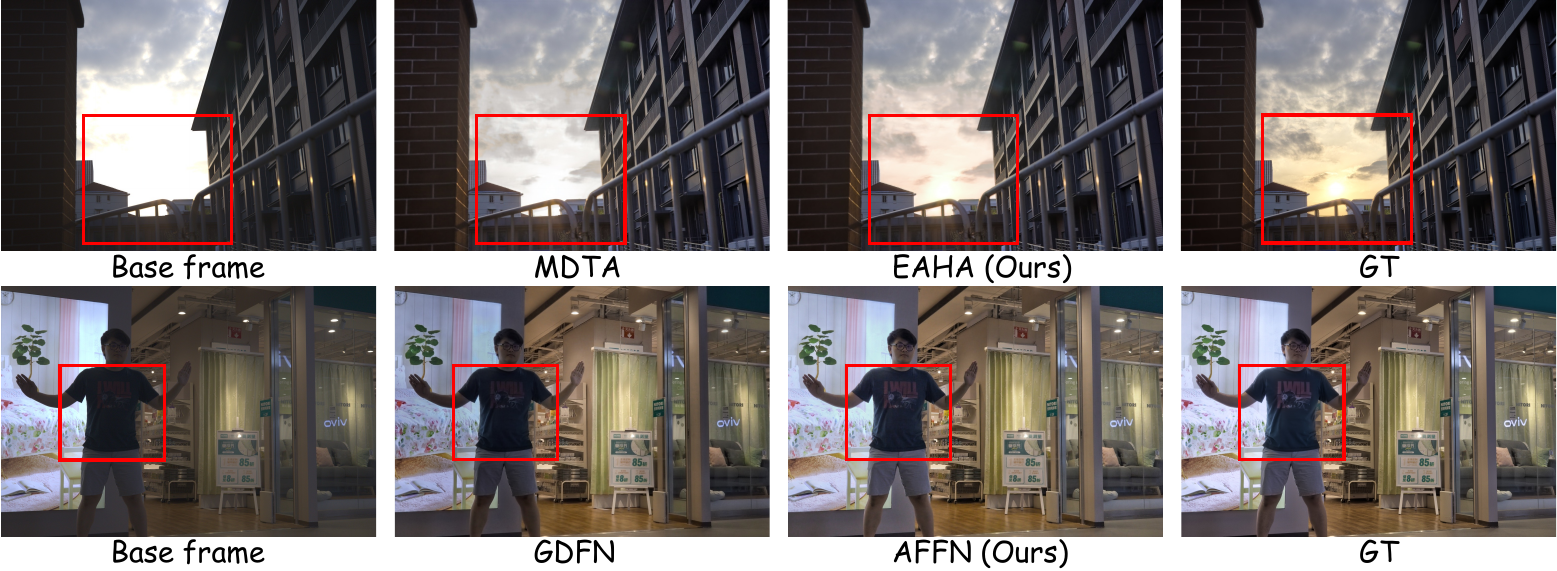}
    \caption{Ablation study of EAHA and AFFN. Their synergy improves restoration in over-exposed regions and the performance of global brightness control.}
    \label{fig:abalation_all}
\end{figure}

\noindent\textbf{Comparison of Fusion Mechanism.} The key difference between our method and previous approaches lies in the fusion mechanism.  We adopt a recurrent fusion strategy based on RSSM, whereas prior methods typically rely on parallel fusion. We analyze the differences between these two approaches at the feature level.  As shown in~\cref{fig:feature_fusion} (a), the conventional parallel fusion method often leads to the introduction of severe ghosting artifacts in the fused features, which the subsequent self-attention and feed-forward networks struggle to mitigate effectively. In contrast, as shown in~\cref{fig:feature_fusion} (b), our pipeline anchors on the base frame and processes global reference information in parallel, which is subsequently fused in a selective manner within the Transformer. This paradigm reduces ghosting artifacts during exposure fusion. To intuitively illustrate the effect of individual modules, we also present qualitative comparisons with and without EAHA and AFFN, as shown in~\cref{fig:abalation_all}.  Benefiting from EAHA's extremity-aware query modulation, our model restores severely over-exposed regions more effectively.
Moreover, through the explicit incorporation of affine modulation, AFFN utilizes multi-frame luminance cues to enable improved global brightness control.

\begin{figure}[t]\small
    \begin{minipage}[t]{0.48\textwidth}
        \centering
        \vspace{-2.5cm}
        \captionsetup{type=table}
        \caption{Quantitative results of different fusion orders on the SICE dataset~\cite{sice_cai2018learning}.}
        \begin{tabular}{c|ccc}
        \midrule
        Fusion Order & PSNR $\uparrow$& SSIM $\uparrow$& LPIPS $\downarrow$\\ \midrule
        (I)   & 22.269 & 0.851 & 0.149 \\
        (II)  & 21.463 & 0.836 & 0.163 \\
        (III) & 22.152 & 0.848 & 0.152 \\
        (IV)  & 22.181 & 0.849 & 0.151 \\ \midrule
        \end{tabular}
        \label{tab:order}
    \end{minipage}
    \hfill 
    \begin{minipage}[t]{0.48\textwidth}
        \centering
        \includegraphics[width=\textwidth]{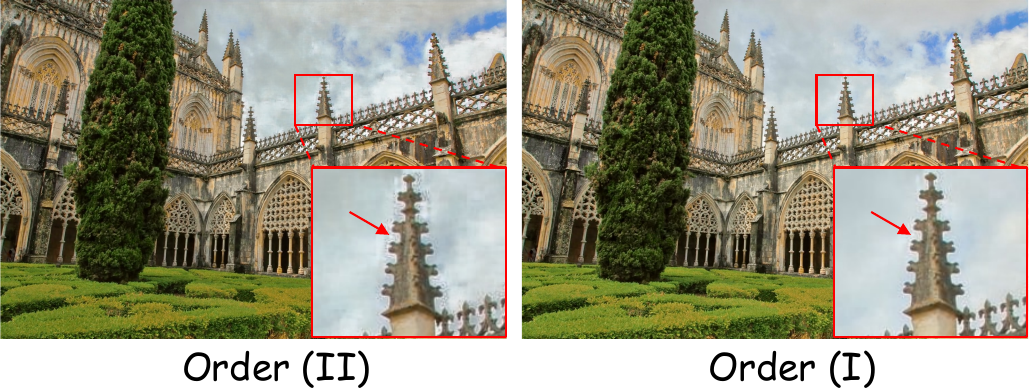}
        \vspace{-4mm}
        \caption{Visual comparison of fusion orders. (II) may introduce some noise.}
        \label{fig:order}
    \end{minipage}
    \vspace{-3mm}
\end{figure}

\noindent\textbf{Fusion Order.} To assess the sensitivity of FreeMEF to fusion order, we perform a comparative study using 5-frame inference on the SICE dataset~\cite{sice_cai2018learning}. Specifically, we evaluate four different fusion strategies: (I) fusing frames in order from the darkest to the brightest exposures (EV-2, EV-1, EV+1, EV+2); (II) fusing from the brightest to the darkest exposures (EV+2, EV+1, EV-1, EV-2); (III) first fusing frames with similar exposures, followed by those with greater exposure gaps (EV-1, EV+1, EV-2, EV+2); and (IV) first fusing frames with the greatest exposure differences, then those with more similar exposures (EV-2, EV+2, EV-1, EV+1). As shown in~\cref{tab:order}, fusion orders (I), (III), and (IV) all achieve relatively robust performance, while order (II) exhibits a significant performance drop. Visualization results in~\cref{fig:order} further reveal that positioning dark frames at the end of the fusion process tends to introduce more noise. Based on these findings, we suggest scheduling the fusion of long-exposure frames at the end of the pipeline to help optimize the final image quality.

\begin{figure}[t]\tiny
    \centering
    \includegraphics[width=\textwidth]{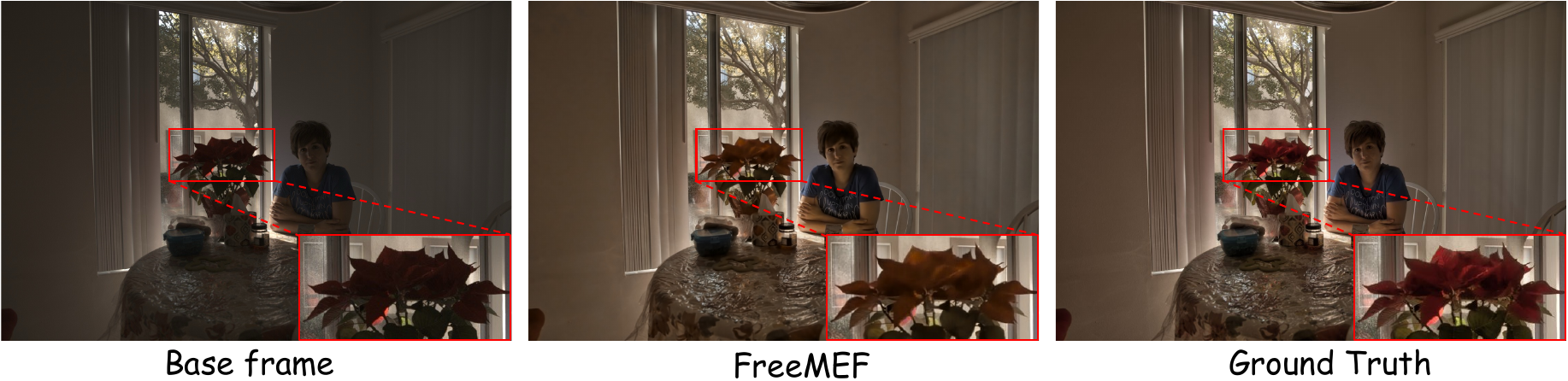}
    \vspace{-5mm}
    \caption{Example of color shift after multi-exposure fusion.}
    \label{fig:limitation}
    \vspace{-2mm}
\end{figure}
\vspace{-2mm}
\section{Conclusion}
\vspace{-2mm}
In this paper, we propose FreeMEF, a flexible and robust transformer framework to tackle the inherent limitations of fixed-input architectures and the similarity paradox in multi-exposure fusion (MEF).  Specifically, to break the constraint of fixed exposure counts, we developed a recurrent state space module (RSSM) that progressively aggregates features from an arbitrary number of input frames. This design endowed the model with the capability to gracefully handle diverse exposure configurations in real-world scenarios without retraining or architectural modification. Furthermore, to overcome the insufficient utilization of reference information in saturated regions, we decoupled the base frame from the reference frames and introduced a global feature guided block (GFGB) for selective fusion.  Within this block, an extremity-aware hybrid attention (EAHA) adaptively balanced cross-attention and self-attention to ensure robust feature retrieval, while an affine-injection feed-forward network (AFFN) dynamically optimized the brightness and contrast of the fused features based on the global context. Extensive experiments demonstrate the effectiveness of our method.

\noindent\textbf{Limitations.} As shown in~\cref{fig:limitation}, the fused result of FreeMEF exhibits slight color deviations compared to the ground truth. This issue primarily stems from overly large gaps between exposure levels, leading to colors that deviate from the true scene appearance.

\noindent\textbf{Acknowledgement.} This work was supported by the Shenzhen Science and Technology Program (No. JCYJ20240813114229039), the Natural Science Foundation of Tianjin, China (No. 24JCZXJC00040), PCL Major Key Project of PCL2025A17-2, the National Natural Science Foundation of China (No. 624B2072), the Doctoral Student Program of the Young S\&T Talents Cultivation Project, CAST, the Fundamental Research Funds for the Central Universities (No. 63263253), the Supercomputing Center of Nankai University (NKSC) and OPPO Research Fund.

\bibliographystyle{splncs04}
\bibliography{main}

\end{document}